\pdfoutput=1

\documentclass[11pt]{article}

\usepackage[preprint]{acl}

\usepackage{times}
\usepackage{latexsym}
\usepackage{graphicx}  
\usepackage{float}  

\usepackage[T1]{fontenc}

\usepackage[utf8]{inputenc}
\usepackage{microtype}

\usepackage{inconsolata}

\usepackage{graphicx}

\usepackage{pifont}
\usepackage{booktabs}

\title{Pretraining Strategies using Monolingual and Parallel Data for Low-Resource Machine Translation}

\author{Idriss Nguepi Nguefack \\
  AIMS Senegal\\
  \texttt{inguepi@aimsammi.org} \\\And
   Mara Finkelstein \\
  Google\\
  \texttt{marafin@google.com}\\\And
  Toadoum Sari Sakayo \\
  AIMS Senegal\\
  \texttt{tsakayo@aimsammi.org}\\}

\begin{document}
\maketitle
\begin{abstract}
This research article examines the effectiveness of various pretraining strategies for developing machine translation models tailored to low-resource languages. Although this work considers several low-resource languages, including Afrikaans, Swahili, and Zulu, the translation model is specifically developed for Lingala, an under-resourced African language, building upon the pretraining approach introduced by \citet{reid2021paradise}, originally designed for high-resource languages. Through a series of comprehensive experiments, we explore different pretraining methodologies, including the integration of multiple languages and the use of both monolingual and parallel data during the pretraining phase. Our findings indicate that pretraining on multiple languages and leveraging both monolingual and parallel data significantly enhance translation quality. This study offers valuable insights into effective pretraining strategies for low-resource machine translation, helping to bridge the performance gap between high-resource and low-resource languages. The results contribute to the broader goal of developing more inclusive and accurate NLP models for marginalized communities and underrepresented populations.  
The code and datasets used in this study are publicly available to facilitate further research and ensure reproducibility, with the exception of certain data that may no longer be accessible due to changes in public availability.\footnote{\url{https://github.com/nguepigit2020/Project1.git}}
\end{abstract}

\section{Introduction}
In recent years, pretraining techniques have gained significant popularity and transformed the field of natural language processing (NLP). Pretraining involves training a language model on a large corpus of text data, enabling it to learn general linguistic patterns and representations. The pretrained model can then be fine-tuned for specific downstream tasks, such as sentiment analysis, named entity recognition, or machine translation, leading to substantial performance improvements. While pretraining has shown remarkable success in high-resource languages like English, Spanish, and Chinese, applying these techniques to low-resource languages poses significant challenges. These languages often lack the extensive datasets required for effective pretraining, limiting the performance of NLP models.

Despite these challenges, researchers have actively explored various pretraining techniques tailored to low-resource languages \cite{costa2022no}. These approaches include unsupervised and semi-supervised learning, transfer learning, and cross-lingual pretraining \cite{khoboko2025optimizing}. Unsupervised methods leverage large amounts of unlabeled data to learn meaningful representations, while semi-supervised techniques enhance model performance by combining limited labeled data with abundant unlabeled data. Transfer learning involves pretraining a model on a high-resource language and fine-tuning it on a low-resource language, exploiting linguistic similarities \cite{zheng2021low}. Cross-lingual pretraining extends this concept by training models on multiple languages simultaneously, enabling them to learn shared representations.

These techniques hold great promise for advancing NLP in low-resource languages, which are often spoken by marginalized and underrepresented communities. Developing more effective pretraining strategies can help mitigate data scarcity and contribute to more accurate, inclusive, and culturally sensitive NLP models \cite{adebara2024cheetah}. Ensuring that NLP technologies benefit all languages and communities, regardless of available resources, is a crucial step toward linguistic inclusivity \cite{okolo2024closing}.

In this work, we evaluate the effectiveness of various pretraining techniques for low-resource languages, with a particular focus on Lingala, an under-resourced African language. To this end, we combined monolingual and parallel data, hypothesizing that this approach would yield the best results. We pretrained multiple models using methods described in \citet{reid20afromt} and \citet{reid2021paradise}, which differ in the types of data used for pretraining. Specifically, we examine the impact of incorporating multiple languages and leveraging both monolingual and parallel data during pretraining. Finally, we evaluated these pretrained models by fine-tuning them on an English-Lingala sequence-to-sequence machine translation task. Our findings offer valuable insights into effective pretraining strategies for low-resource machine translation and contribute to the broader goal of developing more inclusive NLP technologies.

\section{Related Work}
Most previous studies on multilingual pretraining have primarily relied on monolingual data \cite{reid20afromt,pires2019multilingual,song2019mass,liu2020multilingual}. While foundational, this approach does not fully exploit the potential of parallel data. Several proposals have attempted to incorporate parallel data into encoder-only models by training two models simultaneously: one encoder-only model trained on the source language and one decoder-only model trained on the target language. During training, the encoder-only model generates hidden representations of the source sentences, which are then used to train the decoder-only model to generate target sentences \cite{lample2019cross,hu2020explicit}. Some approaches replace words based on a bilingual dictionary, similar to the dictionary denoising objective \cite{wu2019emerging}, while others use multilingual dictionaries but focus only on high-resource languages \cite{reid2021paradise}.

However, these methods often fail to effectively leverage the rich information in parallel data, particularly for low-resource languages. In contrast, sequence-to-sequence models provide a more flexible and natural way to integrate parallel data. Building on the work of \citet{reid2021paradise}, we incorporated parallel corpora into sequence-to-sequence pretraining by feeding concatenated parallel sentences to the encoder and applying different masking strategies. Unlike \citet{reid2021paradise}, our approach specifically targets low-resource languages while maintaining a similar methodology.

Recent advancements in multilingual pretraining have further improved the performance of NLP models for low-resource languages. For example, \cite{pires2019multilingual} demonstrated that multilingual BERT (mBERT) effectively captures cross-lingual representations, though their study primarily focused on high-resource languages. Similarly, \cite{song2019mass} introduced the MASS framework, which employs a masked sequence-to-sequence pretraining objective to enhance machine translation models. However, these studies did not extensively explore the integration of parallel data for low-resource languages.

Additionally, \cite{liu2020multilingual} proposed multilingual denoising pretraining for neural machine translation, highlighting the benefits of training on multiple languages. Their work underscored the importance of leveraging both monolingual and parallel data to improve translation quality. Building on these insights, our study extends these pretraining strategies to low-resource languages.

By integrating both monolingual and parallel data during pretraining, we aim to overcome the limitations of existing approaches and develop more effective strategies for low-resource machine translation. Our work contributes to the growing body of research on multilingual pretraining and provides valuable insights into the development of inclusive and accurate NLP models for underrepresented languages.

\section{Problem}
Despite significant advancements in natural language processing (NLP) and machine translation, the benefits of these technologies are not evenly distributed across all languages. High-resource languages, such as English, Spanish, and Chinese, have seen substantial improvements in translation quality and NLP applications due to the availability of large datasets and extensive research. However, low-resource languages, which are often spoken by marginalized communities and underrepresented populations, continue to lag behind \cite{costa2022no}. This disparity poses significant challenges in various domains, including education, healthcare, and digital communication.

In the context of education, the lack of effective machine translation tools for low-resource languages creates a barrier to accessing educational materials and resources. Students and educators in regions where these languages are spoken often rely on materials in high-resource languages, which can hinder comprehension and learning outcomes. Enhancing machine translation for low-resource languages can facilitate the creation and dissemination of educational content in native languages \cite{steigerwald2022overcoming}, thereby improving educational accessibility and effectiveness.

Moreover, in healthcare settings \cite{al2020implications}, accurate communication is crucial for diagnosing and treating patients. Language barriers can lead to miscommunication, misdiagnosis, and inadequate treatment. Machine translation tools tailored to low-resource languages can help bridge these gaps, ensuring that healthcare providers can effectively communicate with patients who speak these languages.

Additionally, the digital divide is exacerbated by the lack of support for low-resource languages in web-centric applications and technologies \cite{kreienbrinck2024usability}. Users who speak these languages often face difficulties in accessing and interacting with digital content, which limits their participation in the global digital economy. By improving machine translation for low-resource languages, we can make digital platforms more inclusive and accessible to a broader range of users \cite{bella2023towards}.

This research aims to address these challenges by exploring effective pre-training strategies for machine translation models tailored to low-resource languages. Specifically, we focus on Lingala, an under-resourced African language, and investigate the impact of incorporating multiple languages and both monolingual and parallel data during the pretraining phase. Our goal is to develop more accurate and inclusive NLP models that can enhance communication, education, and digital accessibility for speakers of low-resource languages. By doing so, we hope to contribute to the broader goal of reducing linguistic disparities and promoting equitable access to information and services.

\section{Dataset}
We utilized various datasets in our pretraining process for different models, including monolingual datasets (English, Lingala, Afrikaans, Swahili, and Zulu), as well as parallel datasets for fine-tuning. To provide a comprehensive summary of the data used for each language, please refer to Table \ref{data-table} below. It should be noted that we use both monolingual and parallel data for pretraining, while for fine-tuning, only parallel data is used.

\begin{table*}
  \centering
  \begin{tabular}{llcccc}
    \hline
    \textbf{Language} & \textbf{Code} & \multicolumn{2}{c}{\textbf{Parallel data (En-XX)}} & \multicolumn{2}{c}{\textbf{Monolingual data}} \\
    \cmidrule(lr){3-4} \cmidrule(lr){5-6}
    & & \textbf{Size} & \textbf{Sentences} & \textbf{Size} & \textbf{Sentences} \\
    \hline
    Afrikaans & Af & 77MB & 749K & 1.2G & 7979K \\
    Lingala & Ln & 45MB & 388K & 10.3MB & 143K \\
    Swahili & Sw & 80MB & 706K & 2G & 12000K \\
    Zulu & Zu & 75MB & 670K & 18.6MB & 209K \\
    English & En & 77MB & - & 27.2MB & 259K \\
    \hline
  \end{tabular}
  \caption{\label{data-table} Dataset Description}
\end{table*}

\subsection{Data Source}
We used both parallel and monolingual datasets in our study. Specifically, the parallel datasets were obtained from AfroMT \cite{reid20afromt}, a comprehensive benchmark for African language translation. The monolingual datasets Afrikaans, English, Lingala, Swahili, and Zulu were sourced from the open-source CC-100 dataset, which provides a diverse collection of monolingual corpora for various languages. It is worth noting that some of this data may no longer be publicly available. The datasets originate from different sources and vary in size and the number of sentences per language, as detailed in Table \ref{data-table}.

\subsection{Data Quality and Preprocessing}
To ensure the robustness of our models, we applied several preprocessing steps to clean and augment the data. These steps included tokenization, normalization, and duplicate removal. Additionally, we addressed data imbalance, particularly for low-resource languages like Lingala, by employing techniques such as data augmentation and oversampling. These preprocessing steps were essential for improving the quality and consistency of the datasets used in our experiments.

\subsection{Data Split}
Based on the number of sentences per language shown in Table \ref{data-table}, we allocated 3,000 sentences each for testing and validation in the parallel data, with the remaining sentences used for training during both pretraining and fine-tuning. For the monolingual data, we designated 10\% for testing, 10\% for validation, and the remaining 80\% for training in each language, but only during the pretraining phase. This data-splitting strategy ensured a balanced and representative dataset for both the pretraining and fine-tuning phases.

\section{Models and Methods}
\begin{figure*}[t]  
  \centering
  \includegraphics[width=\textwidth]{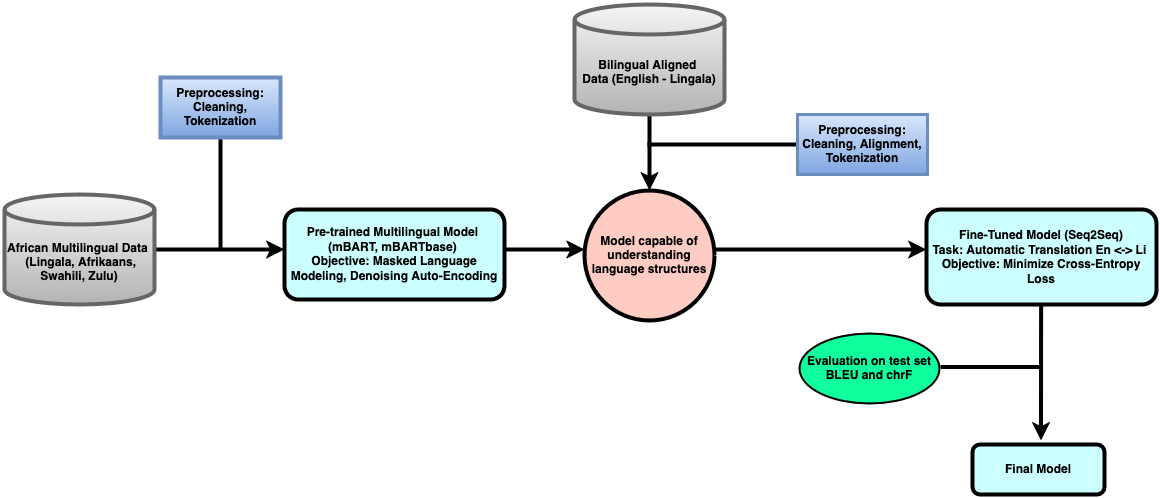}  
  \caption{Flowchart from pretraining to finetuning}
  \label{fig:enter-label}
\end{figure*}
This section presents the models and methodologies used in our study, with a focus on the pretraining and fine-tuning processes (see Figure~\ref{fig:enter-label}). Our goal is to assess the effectiveness of various pretraining techniques and compare their impact on the performance of machine translation models for low-resource languages. Specifically, we pretrain our models on four African languages (Lingala, Afrikaans, Swahili, Zulu) and evaluate them on English-Lingala machine translation tasks.

\subsection{Model Architectures}

\subsubsection{mBART Architecture}
The mBART model \cite{liu2020multilingual} is a sequence-to-sequence denoising autoencoder pretrained on large-scale monolingual corpora spanning 25 languages. It employs a standard Transformer architecture with 12 encoder layers and 12 decoder layers, each featuring a hidden dimension of 1024 and 16 attention heads. The model is trained using a denoising objective, in which the input is corrupted by masking, deleting, or permuting tokens, and the model learns to reconstruct the original sequence. mBART has demonstrated strong performance in multilingual machine translation tasks, particularly in zero-shot and few-shot scenarios.

\subsubsection{AfroBART Architecture}
The AfroBART model \cite{reid20afromt} is a variant of the BART model specifically designed for African languages. It is pretrained on a combination of monolingual and parallel data from eight African languages, including Lingala. AfroBART employs a Transformer architecture similar to mBART but is adapted to the unique characteristics and data constraints of African languages. The model seeks to address the challenges of low-resource languages by leveraging multilingual pretraining and transfer learning techniques.

\section{Hardware and Schedule}
We pre-trained our models on a single machine equipped with two NVIDIA T4 GPUs, 32 vCPUs, and 120 GB of RAM, with each pretraining run taking approximately four days. Fine-tuning was conducted on a machine with one NVIDIA T4 GPU, 32 vCPUs, and 60 GB of RAM, requiring about one day to complete.  

The computational resources used in this study were sufficient to efficiently handle both pretraining and fine-tuning. The NVIDIA T4 GPUs accelerated the training processes, enabling us to run multiple experiments within a reasonable time frame. Additionally, the ample RAM and vCPUs facilitated smooth execution, allowing us to process large datasets and train complex models without significant bottlenecks.  

Pretraining was the most time-intensive phase, requiring up to five days for some experiments, particularly those involving multiple languages and large datasets. In contrast, fine-tuning was relatively faster, taking approximately one day to complete. This efficient use of computational resources and careful time management enabled a thorough evaluation of various pretraining strategies and their impact on machine translation performance for low-resource languages.

\subsection{Pretraining}
To better understand the impact of different data types on the pretraining strategy, we conducted multiple pretraining sessions, making minor modifications such as altering the type of data or the denoising task used. The details of these pretraining approaches are provided in the following sections.

\subsubsection{First Pretraining}\label{first}
For the first experiment, we used only monolingual Lingala data from the AfroMT repository \cite{reid20afromt} (see the dataset description in Table \ref{data-table}). We tokenized all the data using SentencePiece \cite{kudo2018sentencepiece}, employing a multilingual vocabulary of 80k subwords.

We utilized the mBART implementation and the simple denoising task from the fairseq\footnote{\url{https://github.com/facebookresearch/fairseq}\label{link1}} library \cite{ott2019fairseq} to train our models. Our setup included a Transformer-base architecture with a hidden dimension of 768, a feed forward size of 3072, and 6 layers for both the encoder and decoder. The maximum sequence length was set to 1024, and we trained our models with a batch size of 1024 for 100k iterations on a single NVIDIA T4 GPU, with 32 vCPUs and 60 GB of RAM. The training process lasted approximately 24 hours.

\subsubsection{Second Pretraining}\label{second}
For this experiment, we used both the monolingual data of all languages, except English, for the denoising task and the parallel data of all languages for the translation task. This approach combined two tasks (denoising and translation) on two different types of data. While this method was proposed in \cite{reid2021paradise}, it focused primarily on high-resource languages; we applied the same approach but focused on low-resource languages.

Regarding the hyper-parameters, we used SentencePiece \cite{kudo2018sentencepiece} to tokenize all the data, employing a multilingual vocabulary of 80K sub-words. We also used the mBART implementation from the fairseq\ref{link1} \cite{ott2019fairseq} library to train our model. Our configuration included a Transformer-based architecture with a hidden dimension of 768, a feed forward size of 3072, and 6 layers for both the encoder and decoder. The maximum sequence length was set to 1024, and we trained our model with a batch size of 1024 for 100K iterations. The model contained approximately 162 million parameters and the training lasted 4 days.

\subsubsection{Third Pretraining}\label{third}
For this experiment, we used monolingual data from all languages (Afrikaans, English, Lingala, Swahili and Zulu) for the denoising task, as well as parallel data from all languages (English-Lingala, English-Afrikaans, English-Swahili, and English-Zulu) for the translation task. This experiment combined two pretraining tasks (denoising and translation) using two different types of data (monolingual and parallel). It is worth noting that we selected a random sample of 10MB of monolingual data from the English language. By incorporating all languages in this phase, we aimed to achieve a more permanent model.

Regarding the hyper-parameters, we used the same configuration as in the previous experiment. The model had approximately 162 million parameters, and training lasted for 5 days. We later resumed this experiment with the same configuration but increased the English data from 10MB to 112MB.

\subsubsection{Fourth Pretraining}\label{Fourth}
We conducted a fourth experiment in which we used only the parallel and monolingual data of two languages, English and Lingala. It should be noted that we used the same tasks and hyper-parameters as in the previous experiment. Regarding pre-training time, it took about two days.

\subsection{Finetuning}
This section summarizes the various fine-tuning experiments we conducted on existing models, including those we pre-trained ourselves. We utilized the parallel data described in Table \ref{data-table} to fine-tune all pre-trained models, with a primary focus on machine translation between English and Lingala.

As a baseline, we started by fine-tuning the mBART \cite{liu2020multilingual} model, which is pretrained on 25 high-resource languages but does not include Lingala. Next, we fine-tuned the AfroBART \cite{reid20afromt} model, which is pretrained solely on low-resource monolingual data, including Lingala. Finally, we proceeded to fine-tune our own pre-trained models.

We evaluated the system outputs using two automatic evaluation metrics: detokenized BLEU \cite{papineni2002bleu} and chrF \cite{popovic2015chrf}. While BLEU is a standard metric for machine translation, we use chrF to measure performance at the character level, given the morphological richness of the languages in the AfroMT benchmark. Both metrics were calculated using the SacreBLEU library. 

\section{Results and Discussion}
As shown in Table \ref{result-table}, we performed fine-tuning on five pretrained models with highly varied structures. We used the mBART and AfroBART models as starting points to train an automatic translation model from English to Lingala. Evaluation of this model showed that AfroBART outperforms mBART in terms of both BLEU and chrF scores. This can be explained by the fact that the monolingual data used to pretrain AfroBART includes Lingala, the target language of the translation, whereas mBART is pretrained only on high-resource languages.

The BLEU and chrF scores of Experiment 1 are significantly lower than those of mBART and AfroBART. This can be attributed to the fact that the data used to pretrain our experimental model consisted solely of monolingual data from a single language (Lingala), while mBART is pretrained on 25 languages, and AfroBART is trained on 8 African languages, including Lingala. Therefore, we can reasonably conclude that pretraining on multiple languages positively impacts the performance of the resulting translation model.

The scores obtained from the evaluation of the model in Experiment 1 led us to consider another approach that involved using both monolingual and parallel data during the pretraining phase. This approach resulted in a gain of 2 BLEU points and 2 chrF points compared to Experiment 1. Therefore, we can confidently state that the strategy used in this experiment is more beneficial in terms of both BLEU and chrF scores compared to Experiment 1.

We conducted an additional experiment in which we introduced a 10 MB random sample of monolingual English data. This resulted in notably low scores, as shown in Table \ref{result-table}. Subsequently, we increased the size of the English monolingual data to 112 MB, which led to a gain of 4 BLEU points and 3 chrF points.

Finally, we conducted a final experiment using only the parallel and monolingual data of English and Lingala. We observed a significant decrease in both the BLEU and chrF scores, as shown in Table \ref{result-table}. From this, we can affirm that pretraining a model that includes other African languages is more effective than pretraining a model solely on the source and target languages.

\begin{table*}[ht]
  \centering
  \begin{tabular}{lllll}
    \hline
    \textbf{Model} & \textbf{BLEU} & \textbf{chrF} &\textbf{Pretrained Data} & \textbf{Lingala include} \\
    \hline
    mBART & 28.5 & 54.03 & Monolingual & \ding{55} \\
    AfroBART & 29.33 & 54.67 &Monolingual & \ding{51} \\
    Experiment 1 \ref{first} & 25.34 & 51.26 & Monolingual & \ding{51} \\
    Experiment 2 \ref{second} & 27.38 & 53.24 &Monolingual \& Parallel & \ding{51} \\
    Experiment 3 \ref{third} & 21.8 & 48.16 &Monolingual \& Parallel & \ding{51} \\
    Experiment 3$^{+}$ \ref{third} & 25.18 & 51.11 &Monolingual \& Parallel & \ding{51} \\
    Experiment 4 \ref{Fourth} & 21.02 & 48.92 &Monolingual \& Parallel & \ding{51} \\
    \hline
  \end{tabular}
  \caption{\label{result-table} \textbf{Finetuning on top of English and Lingala}}
\end{table*}

\section{Conclusion}
In conclusion, our study aimed to investigate pretraining strategies for machine translation models using low-resource languages. We conducted a series of experiments, gradually introducing monolingual and parallel data to pretrained models.

We first fine-tuned a pretrained model using only parallel data from the source and target languages. Next, we added monolingual data to the pretraining process, which resulted in a significant improvement in the model's BLEU and chrF scores.

Then, we introduced a random sample of English monolingual data, which led to very low scores. However, when we increased the size of the English monolingual data, we observed a notable improvement in the model's translation performance.

Finally, we conducted an experiment using parallel and monolingual data from both English and Lingala. We observed a decrease in BLEU and chrF scores. However, when we pretrained the model using multiple African languages, including the low-resource language, we saw a positive impact on translation performance. Our study underscores the importance of considering pretraining strategies for low-resource languages in machine translation.

It is worth noting that the pretraining approach we used was introduced by \citet{reid2021paradise}, but it originally focused solely on high-resource languages. Our study demonstrates that this approach can also be beneficial for low-resource languages. Interestingly, mBART, which was not pretrained on any African languages, still outperformed our multilingual pretraining setup in some cases. While the scores were close, mBART performed slightly better, which may be attributed to its larger model size, more extensive pretraining on high-quality data, or architectural advantages. This suggests that pretraining on high-resource languages may still offer transferable benefits to low-resource scenarios. Future research in this area can explore different pretraining techniques and incorporate more linguistic knowledge to further improve the performance of machine translation models for low-resource languages.

\section*{Limitations}
Despite the valuable insights gained from this study, there are several limitations to consider. The availability and quality of data for low-resource languages, such as Lingala, significantly impact the effectiveness of the pretraining strategies. Additionally, the findings may not generalize to all low-resource languages due to linguistic differences. Computational resources required for pretraining and fine-tuning can also be prohibitive, and the reliance on BLEU and chrF scores may not fully capture translation quality, especially for morphologically rich languages. Future work should explore more diverse data sources and evaluation methods, such as human evaluation, to better address these challenges.

\section{Acknowledgments}
We would like to extend our heartfelt appreciation to the African Institute for Mathematical Sciences (AIMS) and the African Master's of Machine Intelligence (AMMI) program. Their unwavering support and provision of top-notch machine learning training have been instrumental in the success of our project. We are truly grateful for their guidance and assistance throughout this endeavor.

In addition, we would like to express our deep gratitude to Google for generously granting us access to the Google Cloud Platform (GCP). This invaluable resource enabled us to conduct our experiments with utmost efficiency and effectiveness. We recognize the significant contribution that this grant has made to our research.

Furthermore, we would like to extend a special acknowledgment to the dedicated AMMI staff. Their exceptional assistance and unwavering support have been invaluable to us. Their expertise and commitment have played a crucial role in the development and execution of our project.




\end{document}